\documentstyle[aclap]{article}

\title{Tagset Reduction Without Information Loss}

\author{Thorsten Brants\\
	Universit\"at des Saarlandes\\
	Computerlinguistik\\
	D-66041 Saarbr\"ucken, Germany\\
	{\tt thorsten@coli.uni-sb.de}\\[1ex]
	{\bf In:} {\em Proceedings of ACL-95, Cambridge, MA.}}

\begin{document}

\maketitle


\begin{abstract}

    A technique for reducing a tagset used for $n$-gram part-of-speech
disambiguation is introduced and evaluated in an experiment. The
technique ensures that all information that is provided by the original
tagset can be restored from the reduced one. This is crucial, since we
are interested in the linguistically motivated tags for part-of-speech
disambiguation. The reduced tagset needs fewer parameters for its
statistical model and allows more accurate parameter estimation.
Additionally, there is a slight but not significant improvement of
tagging accuracy.

\end{abstract}


\section{Motivation}

    Statistical part-of-speech disambiguation can be efficiently done
with $n$-gram models \cite{Church88,Cutting92}. These models are
equivalent to Hidden Markov Models ({\em HMMs}) \cite{Rabiner89} of
order $n-1$. The states represent parts of speech ({\em categories,
tags}), there is exactly one state for each category, and each state
outputs words of a particular category. The transition and output
probabilities of the HMM are derived from smoothed frequency counts in a
text corpus.

    Generally, the categories for part-of-speech tagging are
linguistically motivated and do not reflect the probability
distributions or co-occurrence probabilities of words belonging to that
category. It is an implicit assumption for statistical part-of-speech
tagging that words belonging to the same category have similar
probability distributions. But this assumption does not hold in many of
the cases.

    Take for example the word {\em cliff\/} which could be a proper ({\sf
NP})\footnote{All tag names used in this paper are inspired by those
used for the LOB Corpus \cite{Garside87}.} or a common noun ({\sf NN})
(ignoring capitalization of proper nouns for the moment). The two
previous words are a determiner ({\sf AT}) and an adjective ({\sf JJ}).
The probability of {\em cliff\/} being a common noun is the product of the
respective contextual and lexical probabilities $p({\sf NN} | {\sf AT},
{\sf JJ}) \cdot p({\em cliff\/} | {\sf NN})$, regardless of other
information provided by the actual words ({\em a sheer cliff\/} vs.\ {\em
the wise Cliff\/}). Obviously, information useful for probability
estimation is not encoded in the tagset.

    On the other hand, in some cases information {\em not\/} needed for
probability estimation is encoded in the tagset. The distributions for
comparative and superlative forms of adjectives in the Susanne Corpus
\cite{Sampson95} are very similar. The number of correct tag assignments
is {\em not\/} affected when we combine the two categories. However, it
does not suffice to assign the combined tag, if we are interested in the
distinction between comparative and superlative form for further
processing. We have to ensure that the original (interesting) tag can be
restored.

    There are two contradicting requirements. On the one hand, more tags
mean that there is more information about a word at hand, on the other
hand, the more tags, the severer the sparse-data problem is and the larger
the corpora that are needed for training.

    This paper presents a way to modify a given tagset, such that
categories with similar distributions in a corpus are combined without
losing information provided by the original tagset and without losing
accuracy.


\section{Clustering of Tags}

    The aim of the presented method is to reduce a tagset as much as
possible by combining ({\em clustering\/}) two or more tags without
losing information and without losing accuracy. The fewer tags we
have, the less parameters have to be estimated and stored, and the less
severe is the sparse data problem. Incoming text will be disambiguated
with the new reduced tagset, but we ensure that the original tag is
still uniquely identified by the new tag.

    The basic idea is to exploit the fact that some of the categories
have a very similar frequency distribution in a corpus. If we combine
categories with similar distribution characteristics, there should be
only a small change in the tagging result. The main change is that single
tags are replaced by a cluster of tags, from which the original has to
be identified. First experiments with tag clustering showed that, even
for fully automatic identification of the original tag, tagging accuracy
slightly increased when the reduced tagset was used. This might be a
result of having more occurrences per tag for a smaller tagset, and
probability estimates are preciser.

\subsection{Unique Identification of Original Tags}
\label{UniqueIdentSection}

    A crucial property of the reduced tagset is that the original tag
information can be restored from the new tag, since this is the
information we are interested in. The property can be ensured if we
place a constraint on the clustering of tags.

    Let ${\cal W}$ be the set of words, ${\cal C}$ the set of clusters
(i.e.~the reduced tagset), and ${\cal T}$ the original tagset. To restore
the original tag from a combined tag ({\em cluster\/}), we need a unique
function
 \begin{equation}
 \label{RestoreFunction}
	f_{orig}: {\cal W} \times {\cal C} \mapsto {\cal T},
 \end{equation}

    To ensure that there is such a unique function, we prohibit some of
the possible combinations. A cluster is allowed if and only if there is
no word in the lexicon which can have two or more of the original tags
combined in one cluster. Formally, seeing tags as sets of words and
clusters as sets of tags:
 \begin{equation}
 \label{ClusterConstraint}
        \forall c\in{\cal C}, t_1, t_2 \in c,
		t_1 \not= t_2, w \in {\cal W}: ~~~
        w \in t_1 \Rightarrow w \not\in t_2
 \end{equation}
    If this condition holds, then for all words $w$ tagged with a cluster $c$,
exactly one tag $t_{wc}$ fulfills
 \[
        w \in t_{wc} \wedge t_{wc} \in c,
 \]
    yielding
 \[
	f_{orig}(w, c) = t_{wc}.
 \]
    So, the original tag can be restored any time and no information
from the original tagset is lost.

    Example: Assume that no word in the lexicon can be both comparative
({\sf JJR}) and superlative adjective ({\sf JJT}). The categories are
combined to {\sf\{JJR,JJT\}}. When processing a text, the word {\em
easier\/} is tagged as {\sf \{JJR,JJT\}}. Since the lexicon states that
{\em easier\/} can be of category {\sf JJR} but not of category {\sf JJT},
the original tag must be {\sf JJR}.

\subsection{Criteria For Combining Tags}

The are several criteria that can determine the quality of a particular
clustering.

\begin{enumerate}
\item	Compare the trigram probabilities
	$p(B|X_i, A)$, $p(B|A, X_i)$, and $p(X_i|A,B)$, $i = 1,2$.
	Combine two tags $X_1$ and $X_2$, if these probabilities
	coincide to a certain extent.
\item	Maximize the probability that the training corpus is generated
	by the HMM which is described by the trigram probabilities.
\item	Maximize the tagging accuracy for a training corpus.
\end{enumerate}

Criterion (1) establishes the theoretical basis, while criteria (2) and
(3) immediately show the benefit of a particular combination. A measure
of similarity for (1) is currently under investigation. We chose (3) for
our first experiments, since it was the easiest one to implement. The
only additional effort is a separate, previously unused part of the
training corpus for this purpose, the {\em clustering part\/}. We combine
those tags into clusters which give the best results for tagging of the
clustering part.

\subsection{The Algorithm}

    The total number of potential clusterings grows exponential with the
size of the tagset. Since we are interested in the reduction of large
tagsets, a full search regarding all potential clusterings is not
feasible. We compute the local maximum which can be found in polynomial
time with a best-first search.

    We use a slight modification of the algorithm used by
\cite{Stolcke94a} for merging HMMs. Our task is very similar to theirs.
Stolcke and Omohundro start with a first order HMM where every state
represents a single occurrence of a word in a corpus, and the goal is to
maximize the a posteriori probability of the model. We start with a
second order HMM (since we use trigrams) where each state represents a
part of speech, and our goal is to maximize the tagging accuracy for a
corpus.

The clustering algorithm works as follows:
\begin{enumerate}
\item	Compute tagging accuracy for the clustering part with the original
	tagset.
\item	Loop:
	\begin{enumerate}
	\item	Compute a set of candidate clusters (obeying constraint
		(\ref{ClusterConstraint}) mentioned in section
		\ref{UniqueIdentSection}), each consisting of two tags
		from the previous step.
	\item	For each candidate cluster build the resulting tagset
		and compute tagging accuracy for that tagset.
	\item	If tagging accuracy decreases for all
		combinations of tags, break from the loop.
	\item	Add the cluster which maximized the tagging accuracy
		to the tagset and remove the two tags previously used.
	\end{enumerate}
\item	Output the resulting tagset.
\end{enumerate}

\subsection{Application of Tag Clustering}

\begin{table*}
\caption{Tagging results for the test parts in the clustering
        experiments. Exp.\ 1 and 2 are used as the baseline.}
\label{ClusterExpTable}
\begin{center}
\begin{tabular}{|r|ccc|c|}
\hline
   & Training   & Clustering    & Testing       & Result (known words)\\
\hline
1. & parts A and B & --         & part C        & 93.7\% correct \\
2. & parts A and C & --         & part B        & 94.6\% correct \\
\hline
3. & part A     & part B        & part C        & 93.9\% correct \\
4. & part A     & part C        & part B        & 94.7\% correct \\
\hline
\end{tabular}
\end{center}
\end{table*}

    Two standard trigram tagging procedures were performed as the
baseline. Then clustering was performed on the same data and tagging was
done with the reduced tagset. The reduced tagset was only internally
used, the output of the tagger consisted of the original tagset for all
experiments.

    The Susanne Corpus has about 157,000 words and uses 424 tags
(counting tags with indices denoting multi-word lexemes as separate
tags). The tags are based on the LOB tagset \cite{Garside87}.

Three parts are taken from the corpus. Part A consists of about
127,000 words, part B of about 10,000 words, and part C of about 10,000
words. The rest of the corpus, about 10,000 words, is not used for this
experiment. All parts are mutually disjunct.

    First, part A and B were used for training, and part C for testing.
Then, part A and C were used for training, and part B for testing. About
6\% of the words in the test parts did not occur in the training parts,
i.e.\ they are unknown. For the moment we only care about the known words
and not about the unknown words (this is treated as a separate problem).
Table \ref{ClusterExpTable} shows the tagging results for known words.

    Clustering was applied in the next steps. In the third experiment,
part A was used for trigram training, part B for clustering and part C
for testing. In the fourth experiment, part A was used for trigram
training, part C for clustering and part B for testing.

    The baseline experiments used the clustering part for the normal
training procedure to ensure that better performance in the clustering
experiments is not due to information provided by the additional part.

    Clustering reduced the tagset by 33 (third exp.), and 31 (fourth
exp.) tags. The tagging results for the known words are shown in table
\ref{ClusterExpTable}.

    The improvement in the tagging result is too small to be
significant. However, the tagset is reduced, thus also reducing the
number of parameters {\em without\/} losing accuracy. Experiments with
larger texts and more permutations will be performed to get precise
results for the improvement.


\section{Conclusions}

    We have shown a method for reducing a tagset used for
part-of-speech tagging without losing information given by the original
tagset. In a first experiment, we were able to reduce a large tagset and
needed fewer parameters for the $n$-gram model. Additionally, tagging
accuracy slightly increased, but the improvement was not significant.
Further investigation will focus on criteria for cluster selection. Can
we use a similarity measure of probability distributions to identify
optimal clusters? How far can we reduce the tagset without losing
accuracy?


\end{document}